%% file: main.tex
\def\year{2019}\relax
\documentclass[letterpaper]{article} 
\usepackage{aaai19}  
\usepackage{times}  
\usepackage{helvet}  
\usepackage{courier}  
\usepackage{url}  
\usepackage{graphicx}  
\usepackage[utf8]{inputenc} 
\usepackage[T1]{fontenc} 
\usepackage{hyperref}  
\usepackage{booktabs}  
\usepackage{amsfonts}  
\usepackage{nicefrac}  
\usepackage{microtype}  

\usepackage{algpseudocode}
\usepackage[vlined,linesnumbered,ruled]{algorithm2e}

\usepackage{amsmath}
\usepackage{lipsum}
\usepackage{booktabs}
\usepackage{comment}
\usepackage{enumitem}
\usepackage{siunitx}

\algnewcommand\LeftComment[2]{%
\hspace{#1\algindent}$\triangleright$ \eqparbox{COMMENT}{#2} \hfill %
}

\SetKwProg{Fn}{Function}{}{}

\usepackage{tikz}
\usetikzlibrary{decorations.pathreplacing,calc}

\usepackage{floatrow}
\newfloatcommand{capbtabbox}{table}[][\FBwidth]

\begin{document}
\title{HOPF: Higher Order Propagation Framework for Deep Collective Classification}
\author{
  Priyesh Vijayan\textsuperscript{1}\thanks{corresponding author: priyesh@cse.iitm.ac.in}, 
  Yash Chandak\textsuperscript{2}, 
  Mitesh M. Khapra\textsuperscript{1}, 
  Srinivasan Parthasarathy\textsuperscript{3} and 
  Balaraman Ravindran\textsuperscript{1} \\
  \textsuperscript{1}Dept. of CSE and Robert Bosch Centre for Data Science and AI \\
  Indian Institute of Technology Madras, Chennai, India\\
  \textsuperscript{2} Dept. of Computer Science,  University of Massachusetts, Amherst, USA\\ 
  \textsuperscript{3} Dept. of CSE and Dept. of Biomedical Informatics, Ohio State University, Ohio, USA\\
    }
\maketitle

\begin{abstract}
Given a graph where every node has certain attributes associated with it and some nodes have labels associated with them, Collective Classification (CC) is the task of assigning labels to every unlabeled node using information from the node as well as its neighbors. It is often the case that a node is not only influenced by its immediate neighbors but also by higher order neighbors, multiple hops away. Recent state-of-the-art models for CC learn end-to-end differentiable variations of Weisfeiler-Lehman (WL) kernels to aggregate multi-hop neighborhood information. In this work, we propose a Higher Order Propagation Framework, HOPF, which provides an iterative inference mechanism for these powerful differentiable kernels. Such a combination of classical iterative inference mechanism with recent differentiable kernels allows the framework to learn graph convolutional filters that simultaneously exploit the attribute and label information available in the neighborhood. Further, these iterative differentiable kernels can scale to larger hops beyond the memory limitations of existing differentiable kernels. We also show that existing WL kernel-based models suffer from the problem of Node Information Morphing where the information of the node is morphed or overwhelmed by the information of its neighbors when considering multiple hops. To address this, we propose a specific instantiation of HOPF, called the NIP models, which preserves the node information at every propagation step. The iterative formulation of NIP models further helps in incorporating distant hop information concisely as summaries of the inferred labels. We do an extensive evaluation across 11 datasets from different domains. We show that existing CC models do not provide consistent performance across datasets, while the proposed NIP model with iterative inference is more robust. 
\end{abstract}

\input{introduction.tex}

\input{background.tex}

\input{proposed.tex}
\input{results.tex}

\input{conclusion.tex}

\bibliographystyle{aaai}
\bibliography{bibliography.bib}
\input{appendix.tex}

\end{document}

%% file: introduction.tex
\section{Introduction}
Many real-world datasets such as social networks can be modeled using a graph wherein the nodes in the graph represent entities in the network and edges between the nodes capture the interactions between the corresponding entities. Further, every node can have attributes associated with it and some nodes can have known labels associated with them. Given such a graph, collective Classification (CC) 
\cite{neville2000iterative,lu2003link,sen2008collective} is the task of assigning labels to the remaining unlabeled nodes in the graph. A key task here is to extract relational features for every node which not only consider the attributes of the node but also the attributes and labels of its partially labeled neighborhood. Neural network based models have become popular for computing such node representations by aggregating node \& neighborhood information.

The key idea is to exploit the inherent relational structure among the nodes which encodes valuable information about homophily, influence, community structure, etc. \cite{jensen2004collective}. Traditionally, various neighborhood statistics on structural properties \cite{gallagher2010leveraging}, and distributions on labels and features \cite{neville2003collective,lu2003link,mcdowell2013labels} were used as relational features to predict labels. Further, iterative inference techniques were widely adopted to propagate these label predictions until convergence \cite{sen2008collective}. Recently, \cite{kipf2016GCN} proposed Graph Convolutional Networks with a re-parameterized Laplacian based graph kernel (GCN) for the node-level semi-supervised classification task. GraphSAGE \cite{hamilton2017inductive} further extended GCN and proposed few additional neighborhood aggregation functions to achieve state of the art results for inductive learning.

These graph convolution kernels are based on differentiable extensions of the popular Weisfieler-Lehman(WL) kernels. In this work, we first show that a \textit{direct adaptation of WL kernels for CC task is inherently limited as node features get exponentially morphed with neighborhood information} when considering farther hops. More importantly, learning to aggregate information from $K$-hop neighborhood in an end-to-end differentiable manner is not easily scalable. The exponential increase in neighborhood size with increase in hops severely limits the model due to excessive memory and computation requirements. In this work, we propose a Higher-order Propagation framework (HOPF) that provides a solution for both these problems. Our main contributions are:
\begin{itemize}[leftmargin=*]
    \item \textit{A modular graph kernel that generalizes many existing methods}. Through this, we discuss a hitherto unobserved phenomenon which we refer to as \textit{Node Information Morphing}. We discuss its implications on the limitations of existing methods and then discuss a novel family of kernels called the \textit{Node Information Preserving (NIP) kernels} to address these limitations. 
    \item A \textit{hybrid semi-supervised learning framework} for higher order propagation (HOPF) that couples differentiable kernels with an iterative inference procedure to aggregate neighborhood information over farther hops. This allows differentiable kernels to \textit{exploit label information} and further\textit{ overcome excessive memory constraints} imposed by multi-hop information aggregation.
    \item An extensive \textit{experimental study on 11 datasets} from different domains. We demonstrate the NIM issue and show that \textit{the proposed iterative NIP model is robust} and overall outperforms existing models.
\end{itemize}

%% file: background.tex
\section{Background}
  In this section, (i) we define the notations and terminologies used (ii) we present the generic differentiable kernel for capturing higher order information in CC setting (iii) we discuss existing works in the light of the generic kernel and (iv) analyze the Node Information Morphing (NIM) issue.

 \subsection{Definitions and notations} 
    
    Let $G=(V,E)$ denote a graph with a set of vertices, $V$, and edges, $E \subseteq V \times V$. Let $|V| = n$. The set E is represented by an adjacency matrix $A\in \mathbb{R}^{n\times n}$ and let $D\in \mathbb{R}^{n\times n}$ denote the diagonal degree matrix defined as $D_{ii}=\sum_j A_{i,j}$.
    
    A collective classification dataset defined on graph $G$ comprises of a set of labeled nodes, $S$, a set of unlabeled nodes, $U$ with $U = V - S$, a feature matrix: $X \in \mathbb{R}^{n\times f}$ and a label matrix: $Y \in \{0,1\}^{|S|\times l}$, where $f$ and $l$ denote the number of features and labels, respecetively. Let $\hat{Y} \in \mathbb{R}^{n\times l}$ denote the predicted label matrix. 
    
    In this work, neural networks defined over $K$-hop neighborhoods have $K$ aggregation or convolution layers with $d$ dimensions each and whose outputs are denoted by $h_1,\ldots, h_K$. We denote the learnable weights associated with $k$-th layer as $W_k^\phi$ and $W_k^\psi \in \mathbb{R}^{d\times d}$. The weights of the input layer ($W_1^\phi$, $W_1^\psi$) and output layer, $W_L$ are in $\mathbb{R}^{f\times d}$ and $\mathbb{R}^{d\times l}$ respectively. Iterative inference steps are indexed by $t \in (1, T)$. 

\subsection{Generic propagation kernel}
    We define the generic propagation (graph) kernel as follows: 
    \begin{flalign}
        \label{eqn:1} 
        \begin{split}
            h_0 & = X \\
            h_{k} & = \sigma_k (\alpha \cdot(\Phi_k \cdot W_k^\phi) ~~ + ~~ \beta \cdot (F(A)\cdot\Psi_k\cdot W_k^\psi))
        \end{split}
    \end{flalign}
where $\Phi_{k}$ and $\Psi_{k}$ are the node and neighbor features considered at the $k^{th}$ propagation step (layer), $F(A)$ is a function of the adjacency matrix of the graph, and $W_k^\phi$ and $W_k^\psi$ are weights associated with the $k$-th layer of the neural network. One can view the first term in the equation as processing the information of a given node and the second term as processing the neighbors' information. The kernel recursively computes the outputs of the $k$\textsuperscript{th} layer by combining the {\it features} computed till the $(k-1)$\textsuperscript{th} layer. $\sigma_k$ is the activation function of the $k$-th layer and $\alpha$ and $\beta$ can be scalars, vectors or matrices depending on the kernel.  

Label predictions, $\hat{Y}$ can be obtained by projecting $h_K$ onto the label space followed by a sigmoid or softmax layer corresponding to multi-class or multi-label classification task. The weights of the model are learned via backpropagation by minimizing an appropriate classification loss on $\hat{Y}$.

\begin{table*}[t]
\centering

\def\arraystretch{1.25}
\resizebox{\textwidth}{!}{%
  \begin{tabular}{lllllllll}
    \hline
    \multicolumn{1}{|l|}{\textbf{Models}}              & \multicolumn{1}{l|}{\textbf{\begin{tabular}[c]{@{}l@{}}$\pmb{\Phi_k}$ \end{tabular}}} & \multicolumn{1}{l|}{\textbf{F(A)}}           & \multicolumn{1}{l|}{\textbf{\begin{tabular}[c]{@{}l@{}}$\pmb{\Psi_k}$\end{tabular}}} & \multicolumn{1}{l|}{\textbf{$\pmb{\alpha}$}}    & \multicolumn{1}{l|}{\textbf{$\pmb{\beta}$}} & \multicolumn{1}{l|}{\textbf{\begin{tabular}[c]{@{}l@{}}$\pmb{W_k^\phi = W_k^\psi?}$\end{tabular}}}
    & \multicolumn{1}{l|}{\textbf{\begin{tabular}[c]{@{}l@{}} Differentiable \\ Kernel\end{tabular}}} & \multicolumn{1}{l|}{\textbf{\begin{tabular}[c]{@{}l@{}}Iterative \\ Inference\end{tabular}}} \\ \hline
    \multicolumn{1}{|l|}{\textbf{\begin{tabular}[c]{@{}l@{}}BL\_NODE \end{tabular}}}  & \multicolumn{1}{l|}{$h_{0}$}                & \multicolumn{1}{l|}{-}              & \multicolumn{1}{l|}{-}                 & \multicolumn{1}{l|}{1}      & \multicolumn{1}{l|}{-}   & \multicolumn{1}{l|}{-}                 & \multicolumn{1}{l|}{-}                   & \multicolumn{1}{l|}{No}                  \\ 
    \multicolumn{1}{|l|}{\textbf{\begin{tabular}[c]{@{}l@{}}BL\_NEIGH \end{tabular}}} & \multicolumn{1}{l|}{-}                 & \multicolumn{1}{l|}{$D^{-1}A$}        & \multicolumn{1}{l|}{$h_{k-1}$}                 & \multicolumn{1}{l|}{-}      & \multicolumn{1}{l|}{1}   & \multicolumn{1}{l|}{-}                 & \multicolumn{1}{l|}{Yes}                  & \multicolumn{1}{l|}{No}                  \\ 
  \hline
    \multicolumn{1}{|l|}{\textbf{SS-ICA}}               & \multicolumn{1}{l|}{$h_{0}$}                & \multicolumn{1}{l|}{$D^{-1}A$}        & \multicolumn{1}{l|}{$\hat Y$}                 & \multicolumn{1}{l|}{1}      & \multicolumn{1}{l|}{1}   & \multicolumn{1}{l|}{No}                 & \multicolumn{1}{l|}{No}                   & \multicolumn{1}{l|}{Yes}                  \\ 
  \multicolumn{1}{|l|}{\textbf{WL}}               & \multicolumn{1}{l|}{$h_{k-1}$}                & \multicolumn{1}{l|}{$A$}        & \multicolumn{1}{l|}{$h_{k-1}$}                 & \multicolumn{1}{l|}{1}      & \multicolumn{1}{l|}{1}   & \multicolumn{1}{l|}{-}                 & \multicolumn{1}{l|}{-}                   & \multicolumn{1}{l|}{No}                  \\ 

    \multicolumn{1}{|l|}{\textbf{GCN}}              & \multicolumn{1}{l|}{$h_{k-1}$}                & \multicolumn{1}{l|}{$(D+I)^{-1/2}A(D+I)^{-1/2}$} & \multicolumn{1}{l|}{$h_{k-1}$}                 & \multicolumn{1}{l|}{$(D+I)^{-1}$} & \multicolumn{1}{l|}{1}   & \multicolumn{1}{l|}{Yes}                & \multicolumn{1}{l|}{Yes}                  & \multicolumn{1}{l|}{No}                  \\ 

  \multicolumn{1}{|l|}{\textbf{GCN-MEAN}}              & \multicolumn{1}{l|}{$h_{k-1}$}              & \multicolumn{1}{l|}{$D^{-1}A$}       & \multicolumn{1}{l|}{$h_{k-1}$}                 & \multicolumn{1}{l|}{1}      & \multicolumn{1}{l|}{1}   & \multicolumn{1}{l|}{Yes}                 & \multicolumn{1}{l|}{Yes}                  & \multicolumn{1}{l|}{No}                \\ 
  
\multicolumn{1}{|l|}{\textbf{GS-Pool}}              & \multicolumn{1}{l|}{$h_{k-1}$}              & \multicolumn{1}{l|}{$maxpool$}       & \multicolumn{1}{l|}{$h_{k-1}$}                 & \multicolumn{1}{l|}{1}      & \multicolumn{1}{l|}{1}   & \multicolumn{1}{l|}{No}                 & \multicolumn{1}{l|}{Yes}                  & \multicolumn{1}{l|}{No}                \\ 
\multicolumn{1}{|l|}{\textbf{GS-MEAN}}              & \multicolumn{1}{l|}{$h_{k-1}$}              & \multicolumn{1}{l|}{$D^{-1}A$}       & \multicolumn{1}{l|}{$h_{k-1}$}                 & \multicolumn{1}{l|}{1}      & \multicolumn{1}{l|}{1}   & \multicolumn{1}{l|}{No}                 & \multicolumn{1}{l|}{Yes}                  & \multicolumn{1}{l|}{No}                \\ 
    \multicolumn{1}{|l|}{\textbf{GS-LSTM}}               & \multicolumn{1}{l|}{$h_{k-1}$}                & \multicolumn{1}{l|}{LSTM gates}           & \multicolumn{1}{l|}{LSTM}                & \multicolumn{1}{l|}{1}     & \multicolumn{1}{l|}{1}  & \multicolumn{1}{l|}{No}                 & \multicolumn{1}{l|}{Yes}                   & \multicolumn{1}{l|}{No}                  \\
  \hline
    \multicolumn{1}{|l|}{\textbf{NIP-MEAN}}              & \multicolumn{1}{l|}{$h_{0}$}              & \multicolumn{1}{l|}{$D^{-1}A$}       & \multicolumn{1}{l|}{$h_{k-1}$}                 & \multicolumn{1}{l|}{1}      & \multicolumn{1}{l|}{1}   & \multicolumn{1}{l|}{No}                 & \multicolumn{1}{l|}{Yes}                  & \multicolumn{1}{l|}{No}                \\ 
    \multicolumn{1}{|l|}{\textbf{I-NIP-MEAN}}              & \multicolumn{1}{l|}{$h_{0}$}              & \multicolumn{1}{l|}{$D^{-1}A$}       & \multicolumn{1}{l|}{$h_{k-1}, \hat Y$}                 & \multicolumn{1}{l|}{1}      & \multicolumn{1}{l|}{1}   & \multicolumn{1}{l|}{No}                 & \multicolumn{1}{l|}{Yes}                  & \multicolumn{1}{l|}{Yes}    \\ \hline
\end{tabular}}
\caption{ \label{table:hop-inst} Baselines, existing and proposed models seen as instantiations of the proposed framework.}
\end{table*}

 \subsection{Relation to existing works: } 
    Appropriate choice of $\alpha$, $\beta$, $\Phi$, $\Psi$ and $F(A)$ in the generic kernel yield different models. Table \ref{table:hop-inst} lists out the choices for some of the popular models, as well as our proposed approaches. Iterative collective inference techniques, such as the ICA family combine node information with aggregated label summaries of immediate neighbors to make predictions. Aggregation can be based on averaging kernel: $F(A)$=$D^{-1}A$, or label count kernel: $F(A)$=$A$, etc with labels as neighbors features ($\Psi_k$=$\hat{Y}$). This neighborhood information is then propagated iteratively to capture higher order information. ICA also has a semi-supervised variant \cite{mcdowell2012semi} where after each iteration the model is re-learned with updated labels of neighbors. Table: \ref{table:hop-inst} shows how the modular components can be chosen to see semi-supervised ICA (SS-ICA) as a special instantiation of our framework. 
    
    The Weisfeiler-Lehman (WL) family of recursive kernels \cite{weisfeiler1968reduction,shervashidze2011weisfeiler} were initially defined for graph isomorphism tests and most recent CC methods use differentiable extensions of it. In its basic form, it is the simplest instantiation of our generic propagation kernel with no learnable parameters as shown in Table: \ref{table:hop-inst}.
     
    The normalized symmetric Laplacian kernel (GCN) used in \cite{kipf2016GCN} can be seen as an instance of the the generic kernel with node weight, $\alpha$=$(D+I)^{-1}$, individual neighbors' weights' $F(A)$=$(D+I)^{-1/2}A(D+I)^{-1/2}$, $\Phi_k = \Psi_k$ and  $W_k^\phi = W_k^\psi$. We also consider its mean aggregation variant (GCN-MEAN), where $F(A) = D^{-1}A$. In theory, by stacking multiple graph convolutional layers, any higher order information can be captured in a differentiable way in $O(K\times E)$ computations. However in practice, the proposed model in\cite{kipf2016GCN} is only full batch trainable and thus cannot scale to large graph when memory is limited.
    
    GraphSAGE (GS) \cite{hamilton2017inductive} is the recent state-of-the-art for inductive learning. GraphSAGE has also proposed variants of $k^{th}$ order differentiable WL kernels, viz: GS-MEAN, GS-Pool and GS-LSTM. These variants can be viewed as special instances of our generic framework as mentioned in the Table \ref{table:hop-inst}. GS-Pool applies a max-pooling function to aggregate neighborhood information whereas GS-LSTM uses a LSTM to combine neighbors' information sequenced in random order similar to \cite{moore2017DCI}. GS has a mean averaging variant, similar to the to GCN-MEAN model, but treats nodes separately from its neighbors, i.e $W_k^\phi \neq W_k^\psi$. Finally, it either concatenates or adds up the node and neighborhood information. GS-LSTM is over-parameterized for small datasets. With GS-MAX and GS-LSTM there is a loss of information as Max pooling considers only the largest input and LSTM focuses more on the recent neighbors in the random sequence.

 \subsection{Node Information Morphing (NIM): Analysis}
    
In this section, we show that existing models which extract relational features, $h_k$ do not retain the original node information, $h_0$ completely. With multiple propagation steps the $h_0$ is decayed and morphed with neighborhood information. We term this issue as Node Information Morphing (NIM).
    
    For ease of illustration, we demonstrate the NIM issue by ignoring the non-linearity and weights. Based on the commonly observed instantiations of our generic propagation kernel (Eqn: \ref{eqn:1}), where $\Phi_{k} = \Psi_{k} = h_{k-1}$, we consider the following equation:
    \begin{align}
    \label{linear}
    h_{k} &= \alpha*Ih_{k-1} + \beta*F(A)h_{k-1} 
    \end{align}
    
    On unrolling the above expression, one can derive the following binomial form: 
    \begin{align}
        h_{k} &= (\alpha*I + \beta*F(A))h_{k-1} \nonumber \\
        h_{k} &= (\alpha*I + \beta*F(A))^kh_{0} 
        \label{eqn:5}
    \end{align}
    
    From Eqn: \ref{eqn:5}, it can be seen that the relative importance of information associated with node's $0^{th}$ hop information, $h_0$, is $\frac{\alpha^k}{(\alpha + \beta)^k}$. Hence, for any positive $\beta$ the importance of $h_0$ decays exponentially with $k$. It can be seen that the decay rate for GCN is $(D+I)^{-k}$ and $(2)^{-k}$ for the other WL kernel variants mentioned in Table: \ref{table:hop-inst}. 
   
   {\bf Skip connections and Node Information Morphing: \\}
   It can be similarly derived and seen that the information morphing not only happens at $h_0$ but also for every $h_k \forall k \in [0$, $K-1]$. This decay of neighborhood information can be lessened by leveraging skip connections. Consider the propagation kernel in Eqn: \ref{linear} with skip connections as shown below:
   \begin{align}
    \label{eqn:skip}
     h_{k} &= (\alpha*Ih_{k-1} + \beta*F(A)h_{k-1}) + h_{k-1} 
   \end{align}
  The above equation on expanding as above gives:
  \begin{align}
   h_{k} &= ((\alpha + 1)*I + \beta*F(A))^kh_{0} 
  \end{align}

 The relative importance of weights of $h_0$ then becomes $\frac{(\alpha + 1)^k}{(\alpha + \beta + 1)^k}$, which decays slower than $\frac{\alpha^k}{(\alpha + \beta)^k}$ for all $\alpha, \beta > 0$. Though this helps in retaining information longer, it doesn't solve the problem completely. 
 Skip connections were used in GCN to reduce the drop in performance of their model with multiple hops. The addition of skip connection in GCN was originally motivated from the conventional perspective to avoid reduction in performance with increasing neural network layers and not with the intention to address information morphing. In fact, their standard 2 layer model cannot accommodate skip connections because of varying output dimensions of layers. Similarly, GraphSAGE models which utilized concatenation operation to combine node and neighborhood information also lessened the decay effect in comparison to summation based combination models. This can be attributed to the fact that concatenation of information from the previous layer can be perceived as skip connections, as noted by its authors. Though the above analysis is done on a linear propagation model, this insight is applicable to the non-linear models as well. Our empirical results also confirm this.

%% file: proposed.tex
\section{Proposed work}
In this section we propose (i) a solution to the NIM issue and (ii) a generic semi-supervised learning framework for higher order propagation.

 \subsection{Node Information Preserving models}
 To address the NIM issue, we propose a specific class of instantiations of the generic kernel which we call the Node Information Preserving (NIP) models. One way to avoid NIM issue is to explicitly retain the $h_0$ information at every propagation step as in the equation below. This is obtained from Eqn: \ref{eqn:1} by setting $\Phi_k = h_0$ and $\Psi_k = h_{k-1}, \forall k$.    
     \begin{align}
        \label{eqn:GEN-NIP}
        h_k &= \alpha h_{0}W_k^\phi + \beta F(A) h_{k-1}W_k^\psi 
    \end{align}
 
 For different choices of $\alpha, \beta$ and $F(A)$, we get different kernels of this family. In particular, setting $\beta=1-\alpha$ and $F(A)=D^{-1}A$ yields a kernel similar to Random Walk with Restart (RWR) \cite{rwr}. 
    \begin{align}
        \label{eqn:RWR}
        h_k &= \alpha h_{0} + \beta F(A) h_{k-1} 
    \end{align}
 The NIP formulation has two significant advantages: (a) It enables capturing correlation between $k$-hop reachable neighbors and the node explicitly and (b) it creates a direct gradient path to the node information from every layer, thus allowing for better training. 
 We propose a specific instantiation of the generic NIP kernel below:
    \begin{flalign}
    \label{eqn:NIP-MEAN}
    \text{NIP-MEAN}: h_k & = \sigma(\boldsymbol{h_0}W_k^\phi + D^{-1}Ah_{k-1}W_k^\psi) 
    \end{flalign}
    NIP-MEAN is similar to GCN-MEAN but with $\Phi_k=h_0$ and $W_k^\phi \neq W_k^\psi$.
 
 \subsection{Higher Order Propagation Framework: HOPF} 
 Building any end-to-end differentiable model requires all the relational information to be in memory. This hinders models with a large number of parameters and those that process data in large batches. 
 For graphs with high link density and a power law degree distribution, processing even $2^{nd}$ or $3^{rd}$ hop information becomes infeasible. Even with $p$-regular graphs, the memory grows at $O(p^K)$ with the number of hops, $K$. Thus, using a differentiable kernel for even small number of hops over a moderate size graph becomes infeasible. 
 
 To address this critical issue of scalability, we propose a novel Higher Order Propagation Framework (HOPF) which incorporates an iterative mechanism over the differentiable kernels. In each iteration of HOPF, the differentiable kernel computes a $C$ hop neighborhood summary, where $C < K$. Every iteration starts with a summary, ($\Theta^{t-1}$), of the information computed until the $(t-1)$ step as given below. 
    \begin{align}
        \label{eqn:GEN-I-Kernel}
        h_0^0 & = X; ~~~ \Theta^0 = 0  \nonumber\\
        h_k^t & = \sigma(\alpha*\Phi_kW_k^\phi + \beta*F(A)\Psi_k^tW_k^\psi) \\
        \Psi_k^t & = [\Psi_k, \Theta^{t-1}] \nonumber
    \end{align}
 
  After $T$ iterations the model would have incorporated $(K=T \times C)$ hop neighborhood information. Here, we fix $T$ based on the required number of hops we want to capture, $K$, but it can also be based on some convergence criteria on the inferred labels. For the empirical results reported in this work, we have chosen $\Theta^{t-1}$ to be (predicted) labels $\hat{Y}$, along the lines of the ICA family of algorithms. Other choices for $\Theta^{t-1}$ includes the $K$ hop relational information, $h_K$.

   \begin{figure}[!ht]
        \includegraphics[scale=0.33]{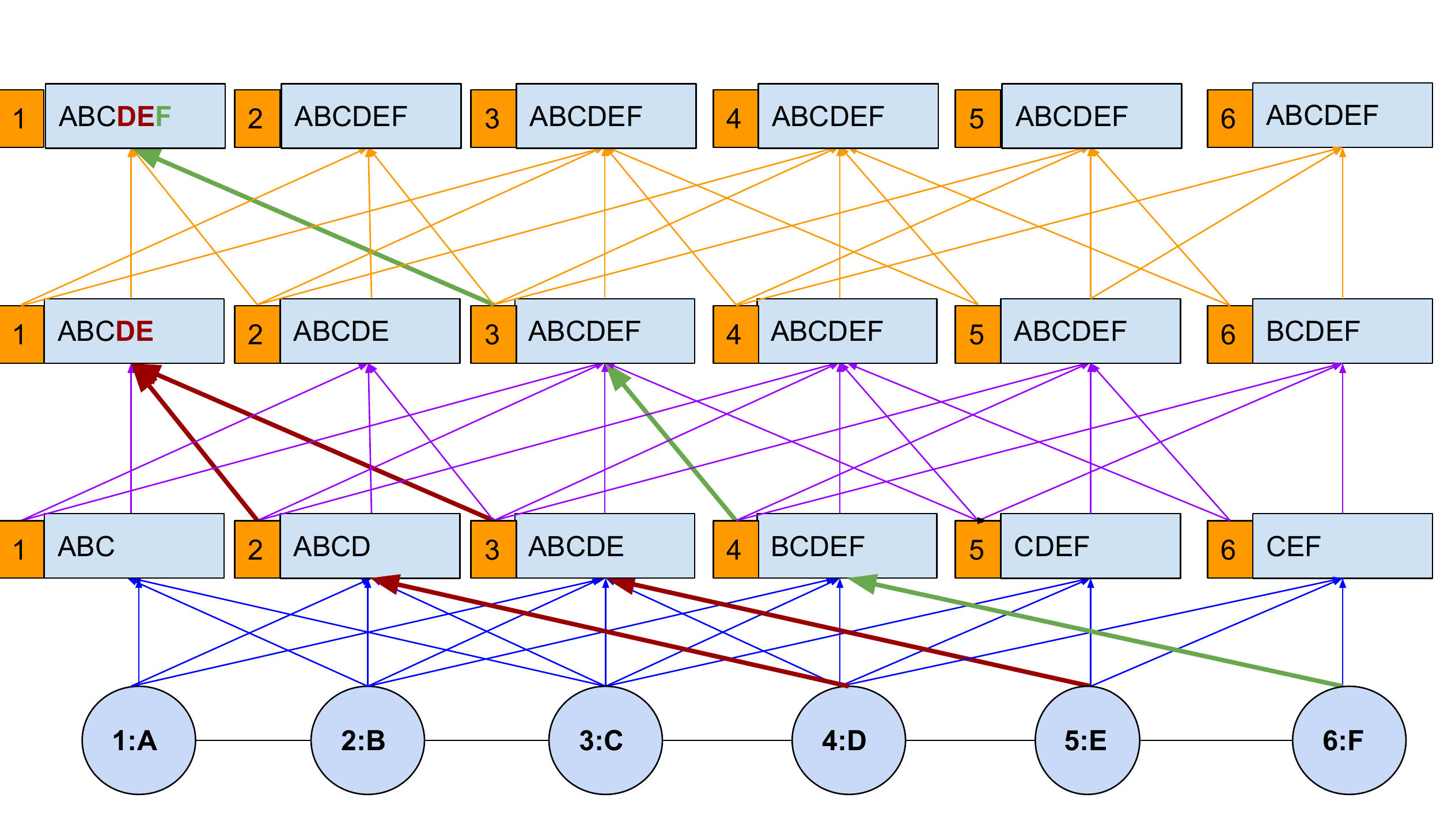}
        \caption{\label{img:chain-example} HOPF explained with a chain graph}
   \end{figure}
 
 We explain HOPF's mechanism with a toy chain graph illustrated in Fig: \ref{img:chain-example}. The graph has 6 nodes with attributes ranging over A-F and the graph kernel used is of the second order. The figure is intended to explain how differentiable and non-differentiable layers are interleaved to allow propagation up to the diameter. We first analyze it with respect to node 1. In the first iteration, node 1 has learned to aggregate attributes from node 2 and 3, viz \textit{BC}, along with its own. This provides it with an aggregate of information from A, B and C. At the start of each subsequent iteration, label predictions are made for all the nodes using a $K^{th}$(In Fig: \ref{img:chain-example}, $K=2$) order differentiable kernel learned in the previous iteration. These labels are concatenated with node attributes to form the features for the current iteration. By treating the labels as non-differentible entities, we
 stop the gradients from propagating to the previous iteration and hence the model is only $K=2$ hop differentiable.
 
 With the concatenated label information, the model can be made to re-learn from scratch or continue on top of the pre-trained model from the last iteration. Following this setup, one can observe that the information of nodes D, E, and F which is not accessible with a $2^{nd}$ order differentiable kernel(blue paths) is now accessible via the non-differentiable paths (red and green paths). In the second iteration, information from nodes at $3^{rd}$ and $4^{th}$ hop (D and E) becomes available and in the subsequent iteration, information from the $5^{th}$ hop (F) becomes available. The paths encoded in blue, purple and orange represent different iterations in the figure and are differentiable only during their ongoing iteration, not as a whole. 

 \begin{algorithm}[!ht]
 	\textbf{\textit{Input:}} { $Dataset: (G, S, U, X, Y)$, \\ No: differentiable hops: $C$, No: of iterations: T} \\
 	\textbf{\textit{Output:}} $\hat{Y}$ \\
 	
 	$\hat{Y}[S]$ = $0$; $\hat{Y}[U]$ = $0$; $\Tilde{Y}$ = $\hat{Y}$  	\\
 	\For {$t$ in 1:T}{
 	    // Learning \\
     	\For {epoch\_id in 1:Max\_Epochs} {
            \For {$nodes$ in $S$}{
                $\tilde{Y}[nodes]$ = predict$(nodes, G, X, \hat{Y}, K)$ \\
                min Loss$(\tilde{Y}[nodes], Y[nodes])$
            }  }
        // Inference \\
        \For {$nodes$ in $U$}{
            $\Tilde{Y}[nodes]$ = predict$(nodes, G, X, \hat{Y}, K)$
        }
        $\hat{Y}[S]$ = $Y$ \\
        // Temporal averaging of predicted labels \\
        $\hat{Y}[U]$= $(T - t)/T *\Tilde{Y} + (t/T)*\hat{Y}[U]$ 
 	}
    \Fn{predict$(nodes, G, X, \hat{Y}, K)$}
    {
        $A, nodes^*$ = get\_subgraph$(G, nodes, K)$ \\
        $X = X[nodes^*]$; $\hat{Y} = \hat{Y}[nodes^*]$ \\
        // Compute 0-hop features\\
        $h_0$ = $\sigma(XW_0)$ \\
        // Compute K-hop features\\
        \For {$k$ in $1:K$}{
              $h_k$ = $\sigma(\alpha \boldsymbol{[h_{0}]}W_k^\phi + \beta F(A, \boldsymbol{[h_{k-1}, \hat{Y}}]W_k^\psi$)) 
        }
        // Predict labels\\
        $\Tilde{Y} = \sigma(h_K[nodes]W_{L})$ \\
        \Return{$\Tilde{Y}$}
     }     
 	\caption{I-NIP-MEAN: Iterative NIP Mean Kernel}
 	\label{Alg:1}  
 \end{algorithm}
 \DecMargin{2em}

\subsection{Iterative NIP Mean Kernel: I-NIP-MEAN}
  
In this section, we propose a special instance of HOPF which addresses the NIM issue with NIP kernels in a scalable fashion. Specifically, we consider the following NIP Kernel instantiation, \textit{I-NIP-MEAN} with mean aggregation function, by setting $F(A)=D^{-1}A$, $\Phi_k=h_0$, $\Psi=h_{k-1}$, $\Theta^{t-1}=\hat{Y}^{t-1}$ and $W_K^\phi \neq W_K^\psi$. 
\begin{align}
\label{eqn:I-NIP}
 h_0^0 & = X; ~~~ \hat{Y}^0 = 0  \nonumber\\
 h_k^{t} & = \sigma(\boldsymbol{h_0^t}W_k^\phi + D^{-1}A[h_{k-1}^{t},\boldsymbol{ \hat{Y}^{t-1}}]W_k^\psi)
\end{align}

In Algorithm \ref{Alg:1} (I-NIP-MEAN), the iterative learning and inference steps are described in \textit{lines: 7-10} and \textit{12-16} respectively. Both learning and inference happen in mini-batches, $nodes$, sampled from the labeled set, $S$ or the unlabeled set, $U$ respectively as shown in $lines: 8$ and $12$ correspondingly. The \textit{predict} function described in \textit{lines:17-27} is used during learning and inference to obtain label predictions for nodes, $nodes$. The procedure first extracts $K$-hop relational features ($h_K$) and then projects it to the label space and applies a sigmoid or a softmax depending on the task, see \textit{line: 27}. 

To extract $K$-hop relational features for $nodes$, the model via \textit{get\_subgraph} function first gathers all $nodes$ along with their neighbors reachable by less than $K+1$ hops ($nodes*$) and represents this entire sub graph by an adjacency matrix ($A$). A $K$-hop representation is then obtained with the kernel as in \textit{lines:21-24}. At each learning phase, the weights of the kernels ($W_k^\phi$s and $W_k^\psi$, $\forall k$) are updated via back-propagation to minimize an appropriate loss function. 

\subsection{Scalability analysis: }


In most real-world graphs exhibiting power law, the size of the neighborhood for each node grows exponentially with the depth of neighborhood being considered. Storing all the node attributes, the edges of the graph, intermediate activations, and all the associated parameters become a critical bottleneck. Here we analyze the efficiency of proposed work to scale to large graphs in terms of the reduction in the number of parameters and space and time complexity. 

\textbf{Number of parameters: } The ratio of available labeled nodes to the unlabeled nodes in a graph is often very small. As observed in \cite{kipf2016GCN,hamilton2017inductive}, the model tends to easily over-fit and perform poorly during test time when additional parameters (layers) are introduced to capture deeper neighborhood. In our proposed framework with iterative learning and inference, the parameters of the kernel at $(t-1)^{th}$ iteration is used to initialize $t^{th}$ kernel and is then discarded, hence the model parameters is $O(C)$ and not $O(K)$. Thus the model can obtain information from any arbitrary hop, $K$ with constant learnable parameters of $O(C)$, where $C=T/K$. But in the inductive setup, the parameter complexity is similar to GCN and GraphSAGE as the kernel parameters from all iterations are required to make predictions for unseen nodes.

\begin{figure}[t]
        \centering
        \includegraphics[scale=0.32]{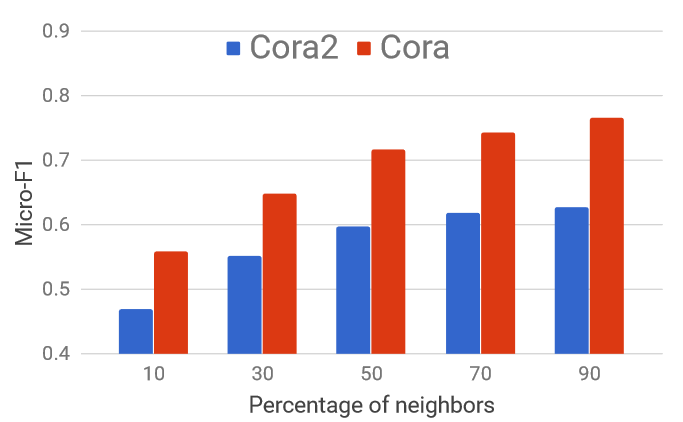}
        \caption{\label{img:2} Impact of NIP-Mean's performance with percentage of neighbors considered}
\end{figure}

\textbf{Space and Time complexity: } For a Graph $G = (V, E)$, we consider aggregating information up to $K$ hop neighborhood. Let number of nodes $N = |V|$, and average degree $p = 2|E|/N$.  For making full batch updates over the graph (like in GCN), computational complexity for information aggregation is $O(NpK)$, and memory required is $O(NK + |E|)$. Even for moderate size graphs, dealing with such memory requirement quickly becomes impractical.  Updating parameters in mini-batches trades off memory requirements with computation time. If batches of size $b$ (where,  $0< b/N << 1$ ) are considered, memory requirement reduces but in worst case, computation complexity increases exponentially to $O(Np^K)$ as neighborhood of size $O(bp^K)$ needs to be aggregated for each of $N/b$ batches independently. In a highly connected graph (such as a PPI, Reddit, Blog etc.), the neighborhood set of a small $K$ may already be the whole network, making the task computationally expensive and often infeasible. o make this tractable, GraphSAGE considers a partial neighborhood information. Though useful, in many cases it can significantly hurt the performance as shown on citation networks in Figure: \ref{img:2}. Not only it hurts the performance but also results in additional hyperparameter tuning for neighborhood size. In comparison, \textit{the proposed work reduces complexity from exponential to linear $O(NTp^C)$} in the total number of hops considered, by doing $T$ iterations of a constant $C$ hop differentiable kernel, such that $T\times C=K$. In our experiments, we found that even $C$ as small as $2$ and $T=5$ was sufficient to outperform existing methods on most of the datasets. The best models were the ones whose $C$ was the largest hop which gave the best performance for the differnetiable kernel. 

\subsection{Miscellaneous related works}

Many extensions of classical methods have been proposed to capture higher-order relational properties of the data. Glocalized kernels \cite{morris2017glocalized} are a variant of the $k$-dimensional Weisfeiler-Lehman \cite{weisfeiler1968reduction} kernel for graph level tasks that use a stochastic approximation to aggregate information from distant nodes. The differentiable kernels are all 1-dim WL-Kernels whose direct adaptation suffers from Node Information Morphing. Relation classifier \cite{macskassy2003simple} builds upon the homophily assumption in the graph structure and diffuses the available label data to predict the labels of unlabelled ones. To make this process more efficient, propagation kernels \cite{neumann2016propagation} provide additional schemes for diffusing the available information across the graph. However, none of these provide a mechanism to adapt to the dataset by learning the aggregation filter.

From a dynamical systems perspective, predictive state representations \cite{sun2016learning} also make use of iterative refinement of internal representations of the model for sequential modeling tasks. However, no extension to graph models has been mentioned.  In computer vision application, iterative Markov random fields \cite{subbanna2014iterative,yu2005combining} have also been shown to be useful for incrementally using the local structure for capturing global statistics. In this work, we restrict our focus to address the limitations of the current state-of-the-art differentiable graph kernels to provide higher order information for collective classification tasks. Moreover, HOPF additionally leverages label information that is found useful. 

Message Passing Neural Network (MPNN) \cite{gilmer2017neural} is a message passing framework which contains the message and read out component. They are defined for graph level tasks. HOPF is explicitly defined for node level tasks and aims at scaling existing graph networks. HOPF's generic propagation kernel is more detailed than MPNN's message component and can additionally support iterative learning and inference. 

%% file: results.tex

\section{Experiments}
\label{Results}
In this section, we describe the datasets used for our experiments, the experimental setup and the models compared. 

\subsection{Dataset details}
\label{Datasets}

We extensive evaluate the proposed models and the baselines on 11  datasets from various domains. In this work, we treat these networks as undirected graphs but the proposed framework can also handle directed graphs with non-negative edges. The datasets used are described below and certain statistics are provided in Table \ref{dataset-stats}. 

\textbf{Social networks:} We use Facebook (FB) from \cite{pfeiffer2015overcoming,moore2017DCI}, BlogCatalog (BLOG) from \cite{wang2010discovering} and Reddit  dataset from \cite{hamilton2017inductive}. In the Facebook dataset, the nodes are Facebook users and the task is to predict the political views of a user given the gender and religious view of the user as features. In the BlogCatalog dataset, the nodes are users of a social blog directory, the user's blog tags are treated as node features and edges correspond to friendship or fan following. The task here is to predict the interests of users. In Reddit, the nodes are the Reddit posts, the features are the averaged glove embeddings of text content in the post and edges are created between posts if the same users comment on both. The task here is to predict the sub-Reddit community to which the post belongs.

\textbf{Citation Networks:} We use four citation graphs: Cora \cite{lu2003link}, Citeseer \cite{bhattacharya2007collective}, Pubmed \cite{namata2012query} and Cora-2 \cite{coraML}. In all the four datasets, the articles are the nodes and the edges denote citations. The bag-of-word representation of the article is used as node attributes. The task is to predict the research area of the article. Apart from Cora-2, which is a multi-label classification dataset from \cite{coraML}, others are multi-class datasets.

\textbf{Biological network:} We use two protein-protein interaction network: Yeast and Human. Yeast dataset is part of the KDD cup 2001 challenge \cite{kdd2001} which contain interactions between proteins. The task is to predict the function of these genes. Similarly, the Human dataset, introduced in \cite{hamilton2017inductive}, is a protein-protein interaction (PPI) network from Human Tissues. The dataset contains PPI from 24 human tissues and the task is to predict the gene's functional ontology. Features consist of positional gene sets, motif gene sets, and immunology signatures.

\textbf{Movie network:} We constructed a movie network from Movielens-2k dataset available as a part of HetRec 2011 workshop \cite{Cantador:RecSys2011}. The dataset is an extension of the MovieLens10M dataset with additional movie tags. The nodes are the movies and edges are created between movies if they share a common actor or director. The movie tags form the movie features. The task here is to predict all possible genres of the movies. 

\textbf{Product network:} We constructed an Amazon DVD co-purchase network which is a subset of Amazon\_060 co-purchase data by  \cite{leskovec2016snap}. The network construction procedure is similar to the one created in \cite{moore2017DCI}. The nodes correspond to DVDs and edges are constructed if two DVDs are co-purchased. The DVD genres are treated as DVD features. The task here is to predict whether a DVD will have Amazon sales rank $\leq$ 7500 or not. To the best of our knowledge there exists no previous work in collective classification that reports results on these many datasets over a wider range of domains. \footnote{Code is available at https://github.com/PriyeshV/HOPF} 
        
\begin{table}
\caption{\label{dataset-stats} Dataset stats: |V|, |E|, |F|, |L|, L$_m$ denote number of nodes, edges, features, labels and is it a multi-label dataset ?} 
\centering
	\begin{tabular}{lllllll}
	    \hline
		Dataset  & Network & |V| & |E| & |F| & |L| & L$_m$ \\ 
		\hline
		Cora  & Citation & 2708 & 5429  & 1433  & 7  & F \\
		Citeseer & Citation & 3312 & 4715 & 3703  & 6  & F  \\
		Cora2 & Citation & 11881 & 34648  & 9568 & 79  & T  \\
		Pubmed & Citation & 19717 & 44327  & 500 & 3  & F  \\
		Yeast & Biology & 1240 & 1674 & 831   & 13  & T  \\
		Human  & Biology & 56944 & 1612348 & 50 & 121  & T   \\
		Reddit & Social & 232965 & 5376619 & 602 & 41  & T  \\
		Blog  & Social & 69814 & 2810844 & 5413 & 46  & T   \\
		Fb & Social & 6302 & 73374  & 2 & 2  & F   \\
		Amazon  & Product & 16553 & 76981 & 30   & 2  & F  \\
		Movie & Movie & 7155 & 388404 & 5297 & 20  & T  \\
		\hline\\
	\end{tabular}
\end{table}

\subsection{Experiment setup:}
The experiments follow a semi-supervised setting with only $10 \%$ labeled data. We consider $20 \%$ of nodes in the graph as test nodes and randomly create 5 sets of training data by sampling $10 \%$ of the nodes from the remaining graph. Further, $20 \%$ of these training nodes are used as the validation set. We do not use the validation set for (re)training. To account for the imbalance in the training set, we use a weighted cross entropy loss (see Appendix) similar to \cite{moore2017DCI} for all the models. In Table: \ref{Results:NIP}, we report the averaged test results for transductive experiments obtained from models trained on the $5$ different training sets. We also report results on the Transfer (Inductive) learning task introduced in \cite{hamilton2017inductive} under their same setting, where the task is to classify proteins in new human tissues (graphs) which are unseen during training.  For detailed information on implementation and hyper-parameter details, kindly refer the Appendix. \\

\begin{table*}[!ht]
\centering
\caption{\label{Results:NIP} Results in Micro-F1 for Transductive experiments. Lower {\it shortfall} is better. Top two results in each column in \underline{bold}.}
\def\arraystretch{1.4}
	\resizebox{\textwidth}{!}{%
\begin{tabular}{|l|lllllllllll|ll|}
\toprule
         & \multicolumn{11}{c|}{Datasets} & \multicolumn{2}{c|}{Aggregate measures} \\ \midrule
MODELS	 & Blog	&FB	&Movie & Cora	&Citeseer	&Cora2  &Pubmed &Yeast  &Human	 &Reddit &Amazon  & Shortfall &  Rank\\
\midrule
BL\_NODE   & 37.929 & \textbf{64.683} & 50.329 & 59.852 & 65.196 & 40.583 & 83.682 & 59.681 & 41.111 & 57.118 & 64.121  & 16.9 & 8.82\\
BL\_NEIGH  & 19.746 & 51.413 & 35.601 & 77.43  & 70.181 & 63.862 & 83.16  & 53.522 & 60.939 & 59.699 & 66.236  & 17.3 & 8.45\\
\hline
GCN        & 34.068 & 50.397 & 39.059 & 76.969 & \textbf{72.991} & \textbf{63.956} & 85.722 & 62.565 & 58.298 & 75.667 & 61.777  & 11.0 & 6.64\\  
GCN-S      & 39.101 & 63.682 & 51.194 & \textbf{77.523} & 71.903 & 63.152 & \textbf{86.432} & 60.34  & 62.057 & 77.637 & \textbf{73.746}  & 4.1 & 4.36 \\
GCN-MEAN   & 38.541 & 62.651 & 51.143 & 76.081 & \textbf{72.357} & 62.842 & 85.792 & 61.787 & 64.662 & 74.324 & 63.674 & 5.6  & 6 \\
GS-MEAN    & \textbf{39.433} & 64.127 & 50.557 & 76.821 & 70.967 & 62.8   & 84.23  & 59.771 & 63.753 & 79.051 & 68.266  & 4.9  & 6\\
GS-MAX   & \textbf{40.275} & 64.571 & 50.569 & 73.272 & 71.39  & 53.476 & 85.087 & 62.727 & 65.068 & 78.203 & 70.302   & 5.5  & 4.73 \\
GS-LSTM    & 37.744 & \textbf{64.619} & 41.261 & 65.73  & 63.788 & 38.617 & 82.577 & 58.353 & 64.231 & 63.169 & 68.024  & 14.4 & 8.45\\
\textbf{NIP-MEAN}    & \textbf{39.433 }& 64.286 & 51.316 & 76.932 & 71.148 & 63.901 & \textbf{86.203} & 61.583 & \textbf{68.688} & 77.262 & 69.136 & 3.6 & 4 \\ 
\hline
SS-ICA  & 38.517 & 64.349 & \textbf{52.433} & 75.342 & 68.973 & 63.098 & 84.798 & \textbf{68.444} & 43.629 & \textbf{81.92}  & 65.789  & 6.6 & 5.73\\
\textbf{I-NIP-MEAN}& 39.398 & 62.889 & \textbf{51.864} & \textbf{78.854} & 71.541 & \textbf{66.23}  & 85.341 & \textbf{69.917} & \textbf{68.652} & \textbf{81.64}  & \textbf{75.045} & \textbf{0.9} & \textbf{2.81} \\
\bottomrule
\end{tabular}}
\end{table*}

\begin{table*}
\centering
\caption{\label{Results:Ind_NIP} Results in Micro-F1 for Inductive learning on Human Tissues}
\def\arraystretch{1.5}
	\resizebox{\textwidth}{!}{%
\begin{tabular}{|l|lllllllllll|}
\toprule
 & Node & Neighbor & NIP-MEAN & GCN-MEAN & GCN  & GS-Mean & GS-Max & GS-LSTM & SS-ICA & I-NIP-MEAN \\ 
\midrule
PPI &44.51	 &83.891	&\textbf{ 92.243 } &86.049	& 88.585	&79.634	 &78.054	 &87.111	&61.51	 &\textbf{92.477} \\
\bottomrule
\end{tabular}}
\end{table*}

\noindent\textbf{Models compared:} 
We compare the proposed NIP and HOPF models with various differentiable WL kernels, Semi-Supervised ICA and two baselines, BL\_NODE and BL\_NEIGH as defined in Table: \ref{table:hop-inst}. BL\_NODE is a K-layer feedforward network that only considers the node's information ignoring the relational information whereas BL\_NEIGH ignores the node's information and considers the neighbors' information. BL\_NEIGH is a powerful baseline which we introduce. It is helpful to understand the usefulness of relational information in datasets. In cases where BL\_NEIGH performs poorer than BL\_NODE, the dataset has less or no useful relational information to extract with the available labeled data and vice versa. In such datasets, we observe no significant gain in considering beyond one or two hops. All the models in Table: \ref{Results:NIP} and Table: \ref{Results:Ind_NIP} except SS-ICA, GCN and GraphSAGE models have skip connections. GraphSAGE models combine node and neighborhood information by concatenation instead of summation.

\subsection{Results and Discussions}
In this section, we make some observations from the results of our experiments as summarized in Tables \ref{Results:NIP} and \ref{Results:Ind_NIP}.

\subsubsection{Statistical significance:} 
In order to report statistical significance of models' performance across different datasets we resort to Friedman's test and Wilcoxon signed rank test as discussed by previous work \cite{demvsar2006statistical}. Levering Friedmans’ test, we can reject the null hypothesis that all the models perform similarly with $p<0.05$. More details and report about the statistical significance of our proposed models, NIP-MEAN and I-NIP-MEAN, over their base variants is presented in the subsequent discussions. .

\subsubsection{Model Consistency}
These rank based significance tests do not provide a metric to measure the robustness of a model across datasets. One popular approach is to use count based statistics like average rank and number of wins. Average Rank of the models across datasets is provided in the table, where lower rank indicates better performance. It is evident from Table \ref{Results:NIP}, that the proposed algorithm I-NIP-MEAN achieves best rank and wins on 4/11 datasets followed by SS-ICA which has 2 wins and NIP-Mean which has 1 win and second best rank. By this simple measure of counting the number of wins of a given algorithm, the proposed method outperforms existing methods. 

However, we argue that this is not helpful at measuring the robustness of models. For example, there could be an algorithm which is consistently the second best algorithm on all the datasets with minute difference from the best and yet have zero wins. To capture this notion of consistency, we introduce a measure, {\it shortfall}, which captures the relative shortfall in performance compared to the best performing model on a given dataset. 
\begin{equation}
\label{shortfall}
\text{\it shortfall[data]} = \frac{\text{\it best[data] - performance[data]}}{\text{\it best[data]}}
\end{equation}
Where {\it best[dataset]} is the micro\_f1 of the best performing model for the dataset and {\it performance[data]} is the model's performance for that dataset. In Table: \ref{Results:NIP}, we report the average {\it shortfall} across datasets. Lower {\it shortfall} indicates a better consistent performance. Even using this measure the proposed algorithm I-NIP-MEAN outperforms existing methods. In particular, notice that while SS-ICA seemed to be the second best algorithm using the naive method of counting the number of wins, it does very poor when we consider the {\it shortfall} metric. This is because SS-ICA is not consistent across datasets and in particular it gives a very poor performance on some datasets which is undesirable. On the other hand, I-NIP-MEAN not only wins on 4/11 datasets but also does consistently well on all the datasets and hence has the lowest average {\it shortfall}. 

\subsubsection{Baselines Vs. Collective classification (CC) models \\}
As mentioned earlier, the baselines BL\_NEIGH and BL\_NODE use \textit{only neighbor} and \textit{only node} information respectively. \textit{In datasets, where BL\_NEIGH significantly outperform BL\_NODE, all CC models ouperform both these baselines by jointly utilizing the node and neighborhood information.} In datasets such as Cora, Citeseer, Cora2, Pubmed and Human, where performance of BL\_NEIGH $>$ BL\_NODE, CC models improve over BL\_NEIGH by up to $8\%$ in the transductive setup. Similarly, on the inductive task where the performance of BL\_NEIGH is greater than BL\_NODE by $\approx 40\%$, CC methods end up further improving by another $8\%$. In Reddit and Amazon datasets, where the performance of BL\_NODE  $\approx$ BL\_NEIGH, CC Methods still learn to exploit useful correlations between them to obtain a further improvement of $\approx 20\%$ and $\approx 10\%$ respectively. 

\subsubsection{WL-Kernels Vs NIP-Kernels}
We make the following observations: 
\hspace{0pt} \newline
\textbf{Node Information Morphing in WL-Kernels: } The poor performance of BL\_NEIGH compared to BL\_NODE on the Blog, FB and Movie datasets suggests that the neighborhood information is noisy and node features are more crucial. \textit{The original GCN which aggregates information from the neighbors but does not use CONCAT or skip connections typically suffers a severe drop in performance of up to $\approx 13\%$ on datasets with high degree.} Despite having the node information, GCN performs worse than BL\_NODE  on these datasets. The improved performance of GCN over BL\_NEIGH in Blog and Movie support that node information is essential. \\
\noindent\textbf{Solving Node Information Morphing with skip connections in WL-Kernels: } The original GCN architecture does not allow for skip connections from $h_0$ to $h_1$ and from $h_{K-1}$ to $h_K$. We modify the original architecture and introduce these skip connections (GCN-S) by extracting $h_0$ features from the $1^{st}$ convolution's node information. \textit{With skip connections, GCN-S outperforms the base GCN on 8/11 datasets.} We observed a performance boost of  $\approx 5-13\%$ in Blog, FB, Movie and Amazon datasets even when we consider only 2 hops thereby decreasing the \textit{shortfall} on these datasets. GCN-S closed the performance gap with BL\_NODE on these datasets and in the case of Amazon dataset, it further improved by another $9\%$. GCN-MEAN which also has skip connections performs quite similarly to GCN-S in all datasets and does not suffer from node information morphing as much as GCN. \textit{It is important to note that skip connections are required not only for going deeper but more importantly, to avoid information morphing even for smaller hops.} 
GS models do not suffer from the node information morphing issue as they concatenate node and neighborhood information. Authors of GS also noted that they observed significant performance boost with the inclusion of CONCAT combination. GS-MEAN's counterpart among the summation models is the GCN-MEAN model which gives similar performance on most datasets, except for Reddit and Amazon where GS-MEAN with concat performs better than GCN-MEAN by $\approx 5\%$. GS-MAX provides very similar performances to GS-MEAN, GCN-MEAN, and GCN-S across the board. Their shortfall performances are also very similar. GS-LSTM typically performs poorly which might be because of the morphing of earlier neighbors' information by more recent neighbors by in the list. \\
\noindent\textbf{Solving Node Information Morphing with NIP Kernels: } \textit{NIP-MEAN, a MEAN pooling kernel from the NIP propagation family outperforms its WL family counterpart, GCN-MEAN on 9/11 datasets. With Wilcoxon signed-rank test, NIP-MEAN > GCN-MEAN with p < 0.01.} It achieves a significant improvement of $\approx 3-6\%$ over GCN-MEAN in Human, Reddit and Amazon datasets. It similarly outperforms GS-MEAN on another 9/11 datasets even though GS-MEAN has twice the number of parameters. NIP-MEAN provides the most consistent performance among the non-iterative models with a shortfall as low as 3.6. NIP-MEAN's clear improvement over its WL-counterparts demonstrates the benefit of using NIP family of kernels which explicitly preserve the node information and mitigate the node information morphing issue. 

\subsubsection{Iterative inference models Vs. Differentiable kernels}
Iterative inference models, SS-ICA and I-NIP-MEAN exploit label information from the neighborhood and scale beyond the memory limits of differentiable kernels. This was evidently visible with our experiments on the large Reddit dataset. Reddit was computationally time-consuming with even partial neighbors due to its high link density. However the iterative models scale beyond 2 hops and consider 5 hops and 10 hops for SS-ICA and I-NIP-MEAN respectively. This is computationally possible because of the linear scaling of time and constant memory complexity of iterative models. Hence, they achieve superior performance with lesser computation time on Reddit. The micro-f1 scores of SS-ICA over iterations on a particular fold was $56.6, 78.4, 79.8, 81.9, 82.2 and 82.2$. Similarly for I-NIP-MEAN on the same fold, we obtained $78, 80.1, 80.7, 81, 81.4 and 81.7$. SS-ICA was remarkable as it can be seen from the table that it managed to obtain 81.92 starting from 57.118 ($BL\_NODE$). 

The benefit of label information over attributes can be analyzed with SS-ICA which aggregates only the label information of immediate neighbors. In Yeast dataset, SS-ICA gains $\approx 8-10\%$ improvement over non-iterative models which do not use label information. However, SS-ICA does not give good performance on some datasets as it does not leverage neighbors features and is restricted to only learn first-order local information unlike multHOPF differentiable WL or NIP kernels.

\subsubsection{Iterative Differentiable kernels Vs. Rest}
\label{sec:stat}
I-NIP-MEAN which is an extension of NIP-MEAN with iterative learning and inference can leverage attribute information and exploit non-linear correlations between the labels and attributes from different hops. \textit{I-NIP-MEAN improves over NIP-MEAN on seven of the eleven datasets with significant boost in performance up to $\approx 3-8\%$ in Cora2, Reddit, Amazon, and Yeast datasets}. Levering Wilcoxon signed-rank test, I-NIP-MEAN is significantly better than NIP-MEAN (with p < 0.05).  I-NIP-MEAN also successfully leverages label information like SS-ICA and obtains similar performance boost on Yeast and Reddit dataset. \textit{It also outperforms SS-ICA on eight of eleven datasets with a statistical significance of p < 0.02 as per the Wlicoxon test} The benefits of using neighbors' attributes along with labels are visible in Amazon and Human datasets where I-NIP-MEAN model achieves $\approx10\%$ and $\approx25\%$ improvement respectively over SS-ICA which uses label information alone.  \textit{Moreover, by leveraging both attributes and labels in a differentiable manner it further achieves a $3\%$ improvement over the second best model in cora2.} \textit{This superior hybrid model, I-NIP-MEAN emerges as the most robust model across all datasets with the lowest shortfall of $\approx 0.9\%$.}

\subsubsection{Inductive learning on Human dataset}
For the inductive learning task in Table: \ref{Results:Ind_NIP}, the cc models obtain a $44\%$ improvement over BL\_NODE by leveraging relational information. The I-NIP-MEAN and NIP-MEAN kernels achieves best performance with a $\approx 6\%$ improvement over GCN-MEAN.  

\subsection{Run time analysis:}

\begin{figure}[t]
        \centering
        \includegraphics[scale=0.40]{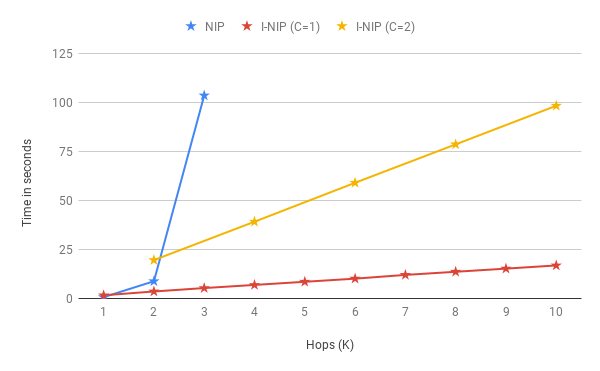}
        \caption{\label{img:3} Iterative models scale linearly}
\end{figure}

We adapt the scalability setup from GCN \cite{kipf2016GCN} to compare average training time per epoch between fully differentiable model, NIP-MEAN and iterative differentible model, I-NIP-MEAN, to make predictions with multHOPF representations. We consider two Iterative NIP-MEAN Variants here: I-NIP-MEAN with 1 differential layer (C=1) and I-NIP-MEAN with 2 differential layerss (C=2). In order to obtain multHOPF representations, we increase the number of iterations, $T$ accordingly. Note: I-NIP-MEAN with C=2 can only provide multHOPF representations of multiples of 2. The training time, included time to pre-fetch neighbors with queues, forward pass of NN, loss computation and backward gradient propagation) similar to the setup of \cite{kipf2016GCN}. We record wall-clock running time for these models to process a synthetic graph with 100k nodes, 500k edges, 100 features, and 10 labels on a 4GB GPU. The batch size and hidden layers size were set to 128.  The plot of the averaged run time over runs across different hops is presented in Figure: \ref{img:3}.

The fully differentiable model, NIP-MEAN incurred an exponential increase in compute time with increase in hop, $(C=K)$ and moreover ran Out-Of-Memory after $3$ hops. Whereas, I-NIP-MEAN with $C=1$ and $C=2$ has a linear growth in compute time with increasing $T$. This is in agreement with the time complexity provided earlier for these models. Not only the time for non-iterative methods increased exponentially with hops, but the memory complexity also increases exponentially with a new layer as it is required to store the gradients and activations for all the new neighbors introduced with a hop. In comparison, the runtime of the proposed iterative solution has a linear growth rate and also has lesser memory footprint.

The choice of K and the decision to use iterative learning depends on a variety of factors such as memory availability  and relevance of the labels. The choice of C, K and T should be determined by performance on a validation set. C should be set as the minimum of maximum differentiable layers that fit into memory or the maximum hop beyond which performance saturates or drops. One should set $T=K/C$, if K doesn’t fit in memory or to an arbitrary constant if performance improves with iteration even if it fits in memory.

%% file: conclusion.tex
\section{Conclusion}
In this work, we proposed HOPF, a novel framework for collective classification that combines differentiable graph kernels with an iterative stage. Deep learning models for relational learning tasks can now leverage HOPF to use complete information from larger neighborhoods without succumbing to over-parameterization and memory constraints. For future work, we can further optimize the framework by committing only high confidence labels , like in cautious ICA \cite{mcdowell2007cautious} to reduce the erroneous information propagation and we can also increase the supervised information flow to unlabeled nodes by incorporating ghost edges \cite{gallagher2008using}. The framework can also be extended for unsupervised tasks by incorporating structural regularization with Laplacian smoothing on the embedding space. 

%% file: appendix.tex
\appendix
\section{Appendix}

\begin{table*}[ht]
	\centering
	\caption{Hyperparameters for different datasets}
	\label{table-hp}
	\resizebox{\textwidth}{!}{
		\begin{tabular}{l|lllllllllll}
			Hyperparams & CORA  & CITE  & CORA2  & YEAST & HUMAN  & BLOG  & FB  & AMAZON & MOVIE & Pubmed & Reddit\\ \hline
			Learning Rate & 1E-02 & 1E-02 & 1E-02  & 1E-02 & 1E-02  & 1E-02 & 1E-02 & 1E-02 & 1E-02 & 1E-02 & 1E-02 \\
			Batch Size & 128  & 128  & 128   & 128  & 512  & 512  & 128  & 512  & 64 & 128 & 512 \\
			Dimensions & 16  & 16  & 128   & 128  & 128  & 128  & 8  & 8  & 128 & 16  & 128 \\
			L2 weight  & 1E-03 & 1E-03 & 1E-06  & 1E-6  & 0  & 1E-06 & 0  & 0  & 1E-06 & 1E-3 & 0\\
			Dropouts  & 0.5  & 0.5  & 0.25  & 0.25  & 0  & 0  & 0  & 0  & 0  & 0.5 & 0 \\
			WCE   & Yes  & Yes  & Yes   & Yes  & No  & Yes  & Yes  & Yes  & Yes & Yes & No\\
			Activation & ReLU  & ReLU  & ReLU  & ReLU & ReLU  & ReLU  & ReLU  & ReLU  & ReLU & ReLU & ReLU \\
			\hline
	\end{tabular}}
\end{table*}

\subsection{Implementation details}
\label{Implementation}
For optimal performance, both iterative and non-iterative models are processed in mini-batches. They make use of queues to pre-fetch the exponential neighborhood information of nodes in a mini-batch. The propagation steps are computed with sparse-dense computations. Mini-batching also makes it possible to efficiently distribute the gradient computation in a multi-GPU setup, we leave this enhancement for future work. The choice of data structure for the kernel is also crucial for processing the graph, i.e trade-off between adjacency list and adjacency matrix results. Working with maxpool or LSTMs are difficult using adjacency matrix as the node's neighborhood information needs to be flattened dynamically. Models based on LSTM will also have to deal with issues of nodes having highly varying degrees and limitations of sequential processing of nodes even at the same hop distance. The code for HOPF framework processes neighborhood information with adjacency matrices and is primarily suited for weighted mean kernels. 

\subsection{Weighted Cross Entropy Loss (WCE)}
\label{WCE}
Models in previous works \cite{yang2016planetoid,kipf2016GCN}, were trained with a balanced labeled set i.e equal number of samples for each label is provided for training. Such assumptions on the availability of training samples and similar label distribution at test time are unrealistic in most scenarios. To test the robustness of CC models in a more realistic set-up, we consider training datasets created by drawing random subsets of nodes from the full ground truth data. 
It is highly likely that randomly drawn training samples will suffer from severe class imbalance. This Imbalance in class distribution can make the weight updates skewed towards the dominant labels during training. To overcome this problem, we generalize the weighted cross entropy defined in \cite{moore2017DCI} to incorporate both multi-class and multi-label setting. We use this as the loss function for all the methods including baselines. The weight $\omega$ for the label $i$ is given in the equation below, where $|L|$ is the total number of labels and $N_j$ represents the number of training samples with label $j$. The weight of each label $\omega_i$ is inversely proportional to the number of samples having that label.

\begin{align}
\label{eqn:wce}
\begin{split}
\omega_i & = \frac{\sum_{j=1}^{|L|}N_j}{|L| \times N_i}
\end{split}
\end{align}

\subsubsection{Hyper-parameters}
The hyper-parameters for the models are the number of layers of neural network (hops), dimensions of the layers, dropouts for all layers and L2 regularization. We train all the models for a maximum of 2000 epochs using Adam \cite{kingma2014adam} with the initial learning rate set to 1e-2. We use a variant of patience method with learning rate annealing for early stopping of the model. Specifically, we train the model for a minimum of 50 epochs and start with a patience of 30 epochs and drop the learning rate and patience by half when the patience runs out (i.e when the validation loss does not reduce within the patience window). We stop the training when the model consecutively loses patience for 2 turns. 

We found all weighted average kernels along with GS-Max model to share similar optimal hyper-parameters as their formulations and parameters were similar. In fact this is in agreement with the work of GCN and GraphSAGE where all their models had similar hyper-parameters. However, GS-LSTM which has more parameters and a different aggregation function required additional hyper-parameter tuning. For reported results, we searched for optimal hyper-parameter setting for a two layer GCN-S model on all datasets with the validation set. We then used the same hyper-parameters across all the other models except for GS-LSTM for which we searched separately. We report performance of models with their ideal number of differentiable graph layers, $K$ based on their performance in validation set. The maximum number of hops beyond which performance saturated or decreased on datasets were: 3 hops for Amazon, 4 hops for Cora2 and HUMAN and 2 hops for the remaining datasets. For the Reddit dataset, we used partial neighbors 25 and 10 in $1^{st}$ and $2^{nd}$ hop which is the default GraphSAGE setting as the dataset had extremely high link density. 

We row-normalize the node features and initialize the weights with \cite{glorot2010understanding}. Since the percentage of different labels in training samples can be significantly skewed, like \cite{moore2017DCI} we weigh the loss for each label inversely proportional to its total fraction as in Eqn: \ref{eqn:wce}. We added all these components to the baseline codes too and ensured that all models have the same setup in terms of the weighted cross entropy loss, the number of layers, dimensions, patience based stopping criteria and dropouts. In fact, we observed an improvement of 25.91 percentage for GraphSage on their dataset. GraphSAGE's LSTM model gave Out of Memory error for Blog, Movielens, and Cora2 as the initial feature size was large and with the large number parameters for the LSTM model the parameter size exploded. Hence, for these datasets alone we reduced the features size.